\documentclass[letterpaper, 10pt, conference]{ieeeconf}
\IEEEoverridecommandlockouts

\usepackage{cite}
\usepackage{amsmath,amssymb,amsfonts}
\usepackage{pgf}
\usepackage{algorithmic}
\usepackage{graphicx}
\usepackage{subfig}
\usepackage{textcomp}
\usepackage{xcolor}
\def\BibTeX{{\rm B\kern-.05em{\sc i\kern-.025em b}\kern-.08em
    T\kern-.1667em\lower.7ex\hbox{E}\kern-.125emX}}
\definecolor{wong-black}        {HTML}{000000}
\definecolor{wong-lightorange}  {HTML}{E69F00}
\definecolor{wong-lightblue}    {HTML}{56B4E9}
\definecolor{wong-green}        {HTML}{009E73}
\definecolor{wong-yellow}       {HTML}{F0E442}
\definecolor{wong-darkblue}     {HTML}{0072B2}
\definecolor{wong-darkorange}   {HTML}{D55E00}
\definecolor{wong-pink}         {HTML}{CC79A7}
\usepackage{csquotes}
\usepackage{booktabs}
\usepackage{hyperref}
\hypersetup{
    colorlinks=true,
    citecolor=wong-green,
    linkcolor=wong-darkblue,
    filecolor=wong-pink,      
    urlcolor=wong-black,
}
\usepackage[T1]{fontenc}
\usepackage{pifont}
\usepackage{makecell}
\usepackage{multirow}
\newcommand{\cmark}{\ding{51}}%
\begin{document}

\title{Mcity Data Engine: Iterative Model Improvement\\Through Open-Vocabulary Data Selection}

\author{Daniel Bogdoll$^{1,2\star}$, Rajanikant Patnaik Ananta$^{1}$, Abeyankar Giridharan$^{1}$,\\Isabel Moore$^{3}$, Gregory Stevens$^{1}$, and Henry X. Liu$^{1}$%
\thanks{$^{1}$ University of Michigan Transportation Research Institute, Ann Arbor, MI, USA
	{\tt\small \{dbogdoll, rpatnaik, abeygiri, gregstvn, henryliu\}@umich.edu}}%
\thanks{$^{2}$ Karlsruhe Institute of Technology, Karlsruhe, BW, Germany}%
\thanks{$^{3}$ Texas A\&M University, College Station, TX, USA}%
\thanks{$^\star$ Work done while Daniel Bogdoll was a research scholar at Mcity of the University of Michigan.}%
}




\maketitle

\begin{abstract}
With an ever-increasing availability of data, it has become more and more challenging to select and label appropriate samples for the training of machine learning models. It is especially difficult to detect long-tail classes of interest in large amounts of unlabeled data. This holds especially true for Intelligent Transportation Systems (ITS), where vehicle fleets and roadside perception systems generate an abundance of raw data. While industrial, proprietary data engines for such iterative data selection and model training processes exist, researchers and the open-source community suffer from a lack of an openly available system. We present the \textit{Mcity Data Engine}, which provides modules for the complete data-based development cycle, beginning at the data acquisition phase and ending at the model deployment stage. The Mcity Data Engine focuses on rare and novel classes through an open-vocabulary data selection process. All code is publicly available on GitHub under an MIT license: \href{https://github.com/mcity/mcity_data_engine}{\textcolor{blue}{https://github.com/mcity/mcity\_data\_engine}}
\end{abstract}

\section{Introduction}
\label{sec:introduction}

\begin{figure}[t!]
    \centering
    \includegraphics[width=1\columnwidth]{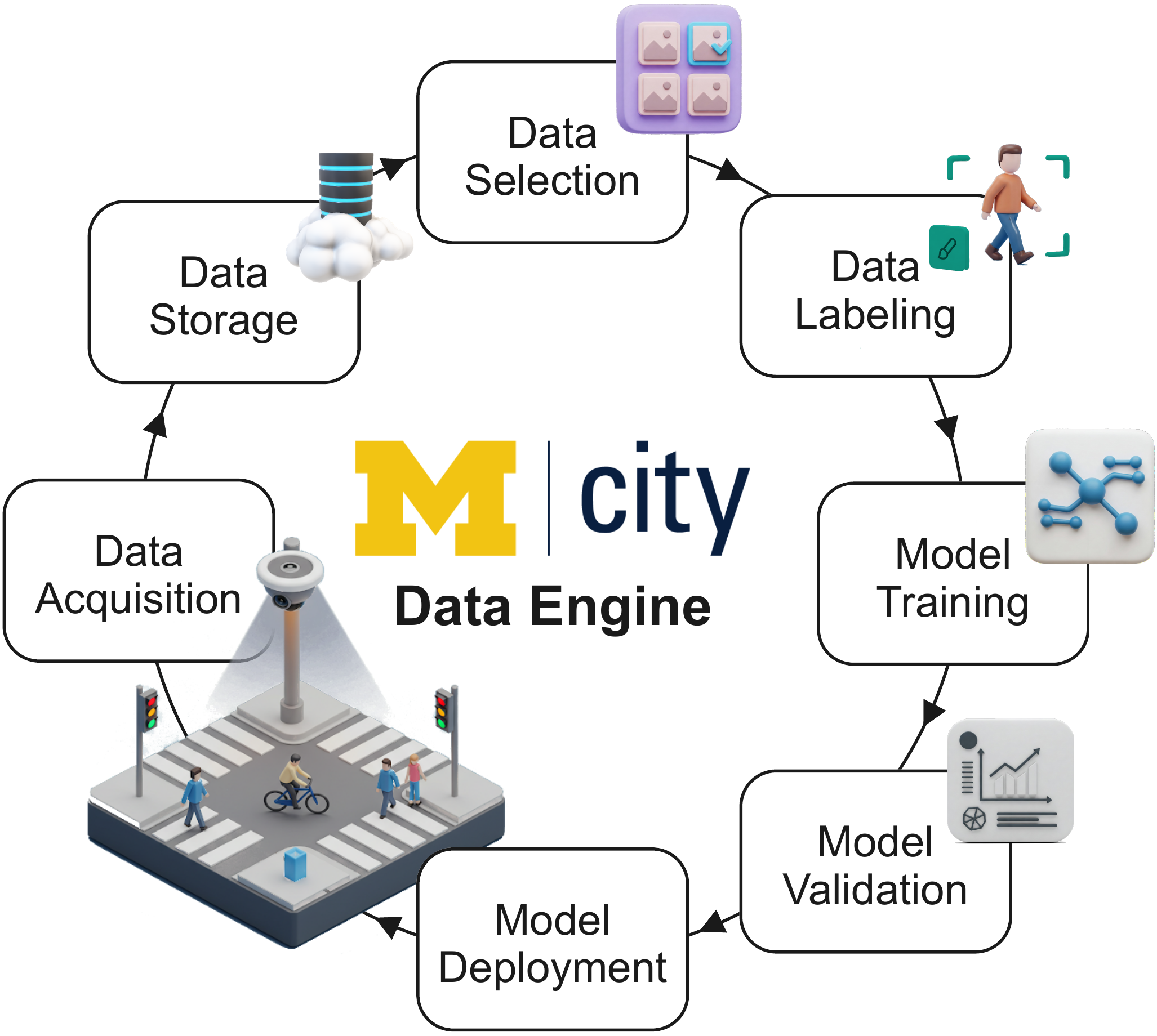}
    \caption{\textbf{Mcity Data Engine:} The open-source Mcity Data Engine provides all modules for data-based development cycles to improve models given unlabeled data and deploy them in real-world scenarios. After data \textit{acquisition} and \textit{storage}, the data engine provides means to \textit{select} relevant samples. These can be \textit{labeled} manually or automatically for model \textit{training}. After a \textit{validation} phase, the model can be \textit{deployed} for inference on the edge or in the cloud.}
    \label{fig:mcity_data_engine}
\end{figure}

\begin{figure*}[t!]
\includegraphics[width=1\textwidth]{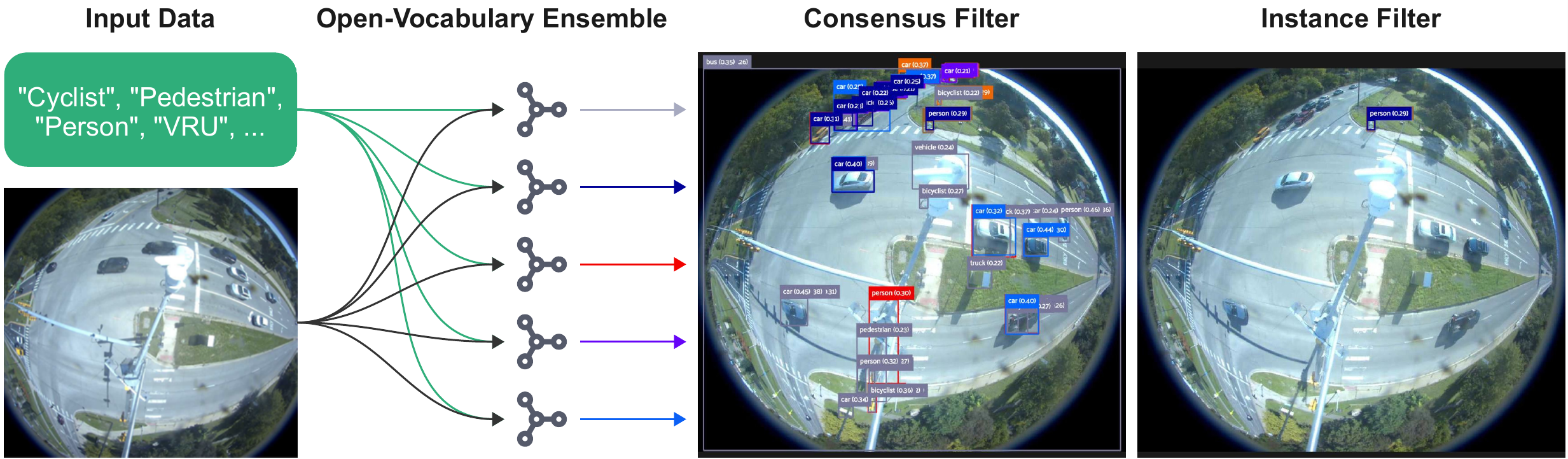}
    \caption{\textbf{Data Selection:} The data selection process supports open-vocabulary data querying based on a list of classes of interest in natural language. An ensemble of task-optimal open-vocabulary object detection models performs inference for a given sample. The resulting bounding boxes are filtered based on a majority consensus decision. Finally, data samples of interest can be selected based on the number of selected instances for subsequent sample-efficient data labeling.}
    \label{fig:data_selection}
\end{figure*}

Deep Learning for intelligent transportation systems has improved significantly over the past years, thanks to the availability of ever-larger datasets~\cite{Bogdoll_Addatasets_2022_VEHITS,liu2024survey}. However, due to introduced domain shifts, these labeled datasets, which originate from a specific source domain, are often insufficient to train models for robust deployments in real-world target domains. To fully leverage the potential of deep learning for intelligent transportation systems in the real world, iterative and data-based development approaches are necessary~\cite{li2023open, ullrichExpandingClassicalVModel2024,Sohn2024whywherewhen}. These require the processing and utilization of large amounts of unlabeled data from the target domain. However, this is challenging due to the long tail of rare occurrences, which are difficult to query. This work presents an iterative approach to leverage an ensemble of open-vocabulary object detection models for a robust selection of rare and novel classes for improved model performance. This method is embedded within our open-source \textbf{Mcity Data Engine} (MDE), which supports the complete data-based development loop from data acquisition to model deployment. Our contributions are twofold:

\begin{itemize}
\item Open-Source Mcity Data Engine provides modules for the complete data-based development cycle (Fig.~\ref{fig:mcity_data_engine})
\item Open-Vocabulary Object Detection Ensemble for data selection of rare and novel classes in raw data (Fig.~\ref{fig:data_selection})
\end{itemize}

\section{Related Work}
\label{sec:related_work}

\begin{table*}[t!]
\resizebox{\textwidth}{!}{
\begin{tabular}{@{}lccccccccc@{}}
\toprule
\textbf{Work}     & \textbf{Year} & \textbf{\makecell{Data \\ Acquisition}} & \textbf{\makecell{Data \\ Storage}} & \textbf{\makecell{Data \\ Selection}} & \textbf{\makecell{Data \\ Labeling}} & \textbf{\makecell{Model \\ Training}} & \textbf{\makecell{Model \\ Validation}} & \textbf{\makecell{Model \\ Deployment}} & \textbf{\makecell{Open \\ Source}} \\ 
\midrule
Nvidia MagLev~\cite{nvidia_maglev}   & 2020 & \cmark                    & \cmark                & \cmark                 & \cmark                 & \cmark                 & \cmark             & \cmark              & --               \\ 
Cruise CLM~\cite{harrisCruisesContinuousLearning2020} & 2020 & \cmark                    & \cmark                & \cmark                 & \cmark                 & \cmark                 & \cmark             & \cmark              & --               \\ 
Tesla Data Engine~\cite{wadatcvprCVPR21WAD2021}   & 2021 & \cmark                    & \cmark                & \cmark                 & \cmark                 & \cmark                 & \cmark             & \cmark              & --               \\ 
\midrule

REM~\cite{jiangImprovingIntraclassLongTail2022a}   & 2022 & --                    & --                & \cmark                 & \cmark                 & \cmark                 & \cmark             & --              & --               \\

SAM Data Engine~\cite{kirillovSegmentAnything2023}   & 2023 & --                    & --                & --                 & \cmark                 & \cmark                 & \cmark             & \cmark              & --               \\

SOLA~\cite{rigoll2023focuschallengesanalysisuserfriendly, Rigoll2024SOLA, rigoll2024clippinglimitsfindingsweet}   & 2024 & --                    & \cmark                & \cmark                 & --                 & --                 & --             & --              & --               \\ 
AIDE~\cite{liang2024aideautomaticdataengine}   & 2024 & --                    & --                & \cmark                 & \cmark                 & \cmark                 & \cmark             & --              & --               \\ 
FullAnno~\cite{hao2024fullannodataengineenhancing,sunImprovingMultimodalLarge2024}   & 2024 & --                    & --                & --                 & \cmark                 & \cmark                 & \cmark             & --              & \cmark               \\ 
ANN~\cite{shoeb2025adaptiveneuralnetworksintelligent}   & 2025 &--                    & --                & \cmark                 & \cmark                 & \cmark                 & \cmark             & --              & --               \\
VLM-C4L~\cite{huVLMC4LContinualCore2025}   & 2025 & --                    & --                & \cmark                 & --                 & \cmark                 & \cmark             & --              & \cmark               \\
VLMine~\cite{yeVLMineLongTailData}   & 2025 & --                    & --                & \cmark                 & --                 & \cmark                 & \cmark             & --              & --               \\

\midrule
\textbf{Mcity Data Engine (ours)}   & 2025 & \cmark                    & \cmark                & \cmark                 & \cmark                 & \cmark                 & \cmark             & \cmark              & \cmark               \\ 
\bottomrule
\end{tabular}
}
\vspace{2mm}
\caption{Overview of Data Engines based on the categories introduced by Li et al.~\cite{liDataCentricEvolutionAutonomous2024a}. The top section shows proprietary industrial data engines. The bottom section shows academic data engines. Our \textbf{Mcity Data Engine} is the first open-source data engine providing functionalities of all categories, comparable to full-fledged industrial solutions.}
\label{tab:sota}
\end{table*}

Large-scale data engines have their roots in autonomous driving, where fleets exist to collect data. Full-fledged solutions like Nvidia's MagLev~\cite{nvidia_maglev}, Cruise's Continuous Learning Machine (CLM)~\cite{harrisCruisesContinuousLearning2020}, or Tesla's Data Engine~\cite{wadatcvprCVPR21WAD2021} are well-known industry examples to leverage large amounts of unlabeled data. However, these systems are proprietary, hindering progress for improved data-based development cycles. More focused academic solutions, however, often consider isolated aspects of data engines, which drastically reduces their utility. Following the characterization of Data Engines by Li et al.~\cite{liDataCentricEvolutionAutonomous2024a}, Table~\ref{tab:sota} provides a comparison of both industrial and academic data engines.

While similar to data engines, pool-based active learning methods struggle with long-tail data mining approaches, as they are primarily designed to reduce labeling cost~\cite{yeVLMineLongTailData} and are thus not the focus of this section. Among the academic contributions to the field of data engines, Jian et al. proposed Rare Example Mining (REM)~\cite{jiangImprovingIntraclassLongTail2022a}. Here, they identify rare objects with a density-based approach. A combination of human and automated labeling is performed, based on which models are iteratively retrained.\\
Kirillov et al.~\cite{kirillovSegmentAnything2023} developed a data engine for the training of the Segment Anything Model (SAM), where they focus on auto-labeling. This way, they use the model to label data and subsequently train the model with the newly labeled data in an iterative fashion. Humans support this approach with manual labels in intermediate stages and quality assurance in later stages. Their final dataset consists of 11 million licensed images. They were able to deploy the model to run in real-time in a web browser.\\
Similarly, Rigoll et al.~\cite{rigoll2023focuschallengesanalysisuserfriendly, Rigoll2024SOLA, rigoll2024clippinglimitsfindingsweet} use Contrastive Language-Image Pretraining (CLIP)~\cite{radfordLearningTransferableVisual2021} combined with panoptic segmentation~\cite{chengMaskedattentionMaskTransformer2022a} to find images and objects based on a text query. Based on a closed-world segmentation, their \textit{search method on the object level for automotive data sets} (SOLA) only works for known classes. They focus on fast data access based on a vector database. Their data selection approach is inspired by prior work leveraging CLIP. Yang et al.~\cite{yangLongTailedObjectMining2022} utilized CLIP to query instances of objects detected by a closed-world object detector. Bogdoll et al. used CLIP to filter out known classes with bounding box proposals coming from detected clusters in LIDAR data, focusing on the unknown~\cite{bogdoll_multimodal_smc_2022}.\\
Liang et al.~\cite{liang2024aideautomaticdataengine} proposed an \textit{Automatically Improving Data Engine} (AIDE), focusing on the automation aspects of data engines by leveraging Vision Language Models (VLM) for the detection of issues. A VLM-generated scene description is compared to a set of known classes to find novel categories, which can suffer from limited expressiveness~\cite{rigoll2023focuschallengesanalysisuserfriendly}. Auto-labeling is performed through open-vocabulary object detection models~\cite{mindererScalingOpenVocabularyObject}, which do not generate individual high-quality detections. Finally, human review is needed for all predictions of novel categories. \\
Hao et al.~\cite{hao2024fullannodataengineenhancing} presented FullAnno, focus on re-labeling data of the existing COCO~\cite{lin2014microsoft} and Visual Genome~\cite{krishna2017visual} datasets. They generate additional bounding boxes and enhance text descriptions. In a related work, Sun et al.~\cite{sunImprovingMultimodalLarge2024} used the re-annotated data engine to train a Vision Language Model.\\
Shoeb et al.~\cite{shoeb2025adaptiveneuralnetworksintelligent} focus on the integration of previously unknown classes into the training process of neural networks. They focus on an \textit{Adaptive Neural Network} (ANN) architecture to integrate new heads for novel classes. Their approach requires annotated out-of-distribution (OOD) objects for the detection of further unknowns and relies on a human classification or selection of detected OOD instances for further training. While proposing a full-scale data engine from data collection to deployment, they demonstrate their work only on publicly available datasets~\cite{cordts2016cityscapes,maag2022two}.\\ 
Hu et al. presented VLM-C4L~\cite{huVLMC4LContinualCore2025}. Given a labeled dataset with corner cases, they use VLMs to create subsets of interest. By combining a core dataset with corner case data and iteratively fine-tuning a model, it adapts to new scenarios. They focus on visually challenging scenarios.\\
Ye et al. proposed VLMine~\cite{yeVLMineLongTailData}, where they use a Vision Language Model~\cite{liuVisualInstructionTuning} to describe images and use a keyword-frequency-based approach based on a Large Language Model (LLM) to select long-tail samples. These are then used to retrain models iteratively. While they propose to label newly found instances, their work is based on the labeled Waymo Open Dataset~\cite{sun2020scalability}.

Most academic works only contribute parts to the concept of a holistic data engine, and many do not provide accompanying source code, reducing the utility of their works. A common trend is the utilization of Vision Language Models for data mining, but current works do not focus on the robustness of those predictions. To the best of our knowledge, the Mcity Data Engine is the first open-source Data Engine of its kind for holistic data-based development cycles, comparable to full-fledged industrial solutions. It leverages a robust ensemble of open-vocabulary object detection models for a high-quality data selection process based on queries formulated in natural language.

\section{Method}
\label{sec:method}

The Mcity Data Engine is a framework for data-based development cycles. It provides modules for all stages of the process, starting with data \textit{acquisition} and \textit{storage}. The MDE is domain-agnostic and supports vision-based datasets from a wide array of sources, performing all actions on a unified dataset format~\cite{moore2020fiftyone}. As of May 01, 2025, the Mcity Data Engine is publicly available under version 1.1~\cite{bogdoll2025mcitydataengine}.

It addresses the needs of real-world practitioners, who are often confronted with enormous amounts of unlabeled data and struggle to decide which data to label and use for training. It is especially well-suited for the detection of rare and novel long-tail classes. In the following, the individual modules of the Mcity Data Engine will be presented.

\subsection{Data Acquisition and Storage}

The MDE supports a wide variety of data sources, focusing on vision datasets. These sources can either be \textit{dynamic}, such as direct access to real-world cameras, or \textit{static}, such as in the form of datasets. In both cases, data is converted into the open-source Voxel51 format~\cite{moore2020fiftyone} with all metadata stored in a dedicated MongoDB~\cite{mongo_db}. For dynamic data sources, such as camera data from the Smart Intersections Project (SIP)~\cite{sip_umtri} in Ann Arbor, Michigan, USA, the Mcity Data Engine provides an access scheme to utilize data in AWS S3 buckets~\cite{aws_s3} as a data storage system, providing native date range filtering and optional downsampling if the temporal context is of less relevance.
The MDE utilizes individual import modules for static data sources, such as the Fisheye8K dataset~\cite{gochoo2023fisheye8kbenchmarkdatasetfisheye}, to load locally stored data into a common format. This approach is fully generic and extendable to any visual dataset, with multiple import modules provided.


\subsection{Open-Vocabulary Data Selection}

As shown in Fig.~\ref{fig:data_selection}, the Mcity Data Engine leverages an ensemble of open-vocabulary models -- which are sometimes also called zero-shot models -- for the selection of samples with classes of interest. To leverage strengths from different models, the MDE utilizes four different architectures: OWL-ViT~\cite{mindererSimpleOpenVocabularyObject}, OWLv2~\cite{mindererScalingOpenVocabularyObject}, Grounding-DINO~\cite{liuGroundingDINOMarrying2025}, and OmDet-Turbo~\cite{zhao2024real, zhao2024omdet}. Since there are a total of twelve different sizes and variants of the models, it is advised to utilize a task-optimized subset during the data selection process for increased efficiency. This can be performed based on a small labeled seed dataset, as demonstrated in Sec.~\ref{sec:eval}.

To select data of interest, users of the Mcity Data Engine simply provide a list of classes of interest in natural language, such as $C_I = \{person, pedestrian, cyclist\}$. A task-optimized ensemble of open-vocabulary models performs inference on the data of interest, generating bounding boxes for all requested classes that were detected. This open-world approach provides much higher flexibility compared to closed-world approaches, which have to focus on atypical or rare instances of known classes~\cite{yangLongTailedObjectMining2022,Rigoll2024SOLA}. As the prediction quality of current open-vocabulary models is still suboptimal~\cite{mindererScalingOpenVocabularyObject}, a consensus-based filter stage takes place next to eliminate a majority of noise in the form of false positives and false negatives. Model consensus is based on an Intersection over Union (IoU) threshold. As shown in Fig.~\ref{fig:data_selection}, this filter requires a subset of $C_I$, only including the classes of interest. Other classes, such as $\{car, vehicle\}$ could be used in an additional stage to improve filtering, which was not deemed necessary. Finally, samples can be selected based on the number of detections per frame. This allows for the selection of frames with many instances, which can reduce the number of samples to label while maintaining a higher number of instances of interest. The MDE provides parallelized multi-GPU inference, as most current open-vocabulary models are still slow. With a typical setting of 8 $\times$ Nvidia H100 nodes, the MDE processes over 1,300,000 high-resolution samples per day.


The Mcity Data Engine provides further machine learning-based data selection methods utilizing sample embeddings~\cite{griffinZeroShotCoresetSelection2024, radfordLearningTransferableVisual2021} and anomaly detection techniques~\cite{akcay2022anomalib,Bogdoll_Anomaly_2022_CVPR}. We refer interested readers to the official GitHub repository~\cite{bogdoll2025mcitydataengine} for further details.

\subsection{Data Labeling and Model Training}

Once the data selection process is finalized, data of interest can be labeled. The Mcity Data Engine supports both manual and automatic labeling. For human inspection and labeling, the MDE integrates with the open-source Computer Vision Annotation Tool (CVAT)~\cite{cvat,moore2020fiftyone}. If a labeled seed dataset exists, the newly labeled samples can be merged with the existing labeled data. For auto-labeling, multiple modes exist. First, detections from the utilized open-vocabulary models~\cite{mindererScalingOpenVocabularyObject, mindererSimpleOpenVocabularyObject, zhao2024real, liuGroundingDINOMarrying2025} can be utilized. Second, pre-trained models for segmentation~\cite{kirillovSegmentAnything2023} and depth estimation~\cite{yangDepthAnythingUnleashing2024, bochkovskiiDepthProSharp2024,bhatZoeDepthZeroshotTransfer2023} can generate pixel-wise masks. Finally, previously trained models can be utilized for auto-labeling, which is especially interesting for slow but high-quality models.

In an intermediate stage prior to training, the MDE provides a class alignment process for data with similar classes but different label schemes. For example, a dataset $A$ only provides a class set $C_A=\{person\}$, but a dataset $B$ provides more granular classes $C_B=\{pedestrian, cyclist\}$. Based on zero-shot classification models~\cite{radfordLearningTransferableVisual2021,zhaiSigmoidLossLanguage2023,tschannenSigLIP2Multilingual2025}, the Mcity Data Engine first quantifies how alike the classes in different datasets are using a similarity score between 0 and 1, represented as $z:C_A \times C_{B} \rightarrow [0,1]$. Then, it uses these similarity scores to perform an upscaling process, $U(z):C_A \rightarrow C_{B}$, which effectively maps the broader labels of dataset $A$ to the more detailed labels of dataset $B$, thus aligning their label schemes. Based on a threshold, the original class can be retained if no fit is detected. Given labeled data, the MDE supports three different model sources for training: HuggingFace Transformers~\cite{wolf-etal-2020-transformers}, Ultralytics~\cite{Jocher_Ultralytics_YOLO_2023}, and individual models, such as Co-DETR~\cite{zong2023detrs} as part of the MMDetection~\cite{mmdetection} family, through containerized access via Docker or Singularity. The MDE automatically generates train, val, and test splits if necessary. As of May 01, 2025, 22 model architectures for the task of object detection are available through the MDE for training. This includes SotA model architectures such as DAB-DETR~\cite{liuDABDETRDynamicAnchor2021}, DETA~\cite{ouyang-zhangNMSStrikesBack2022}, Co-DETR~\cite{zong2023detrs}, RT-DETRv2~\cite{lvRTDETRv2ImprovedBaseline2024}, and YOLOv12~\cite{tianYOLOv12AttentionCentricRealTime2025}.

\subsection{Model Validation and Deployment}

Through the unified Voxel51~\cite{moore2020fiftyone} dataset representation, a full model evaluation suite is provided. This includes all standard metrics, such as Precision, Recall, and F1 Score, and allows for the direct comparison between models.
The Mcity Data Engine supports model deployment for real-world use cases on the edge. For several models integrated into the MDE~\cite{Jocher_Ultralytics_YOLO_2023}, an export module for the serialized Open Neural Network Exchange (ONNX) format is provided.

The Mcity Data Engine is compatible with the real-world Msight roadside perception system~\cite{zhangDesignImplementationEvaluation2022,zouRealtimeFullstackTraffic2022,zhangMSightEdgeCloudInfrastructurebased2023} as used in Ann Arbor, Michigan, USA. Here, deployed models are used to identify traffic participants in various locations, enabling tasks such as near-miss detection~\cite{annarbor_nearmiss,tottel2021reliving}. This compatibility is achieved through a well-defined model interface and is extensible to other real-world systems with different Operational Design Domains (ODD).
\section{Evaluation}
\label{sec:eval}

We evaluate and demonstrate the utility of the Mcity Data Engine with a real-world use case of Vulnerable Road User~(VRU) detection in fisheye camera data, as shown in Fig.~\ref{fig:data_selection}. Based on large amounts of unlabeled camera data through the Smart Intersections Project in Ann Arbor, Michigan, USA, we are interested in improving object detection for the VRU target classes $C_T = \{pedestrian, cyclist\}$. Under this evaluation setting, we leverage the MDE to detect new data of interest in the SIP data stream, label these samples, and retrain the model under test, all in an efficient manner.

\subsection{Selection of Open-Vocabulary Models}

Given a large amount of unlabeled data from the SIP project, we leverage open-vocabulary models for data selection. For a task-optimized selection to form an ensemble of pre-trained models, we evaluate a selection of twelve state-of-the-art~(SotA) models from four different model architectures. Model performance is evaluated based on the seed dataset $\mathcal{D}_{seed}$ with 1,260 instances of VRUs in the form of 1,179 pedestrians and 81 cyclists, as shown in Table~\ref{tab:open_vocab_eval}.

\begin{table}[h]
\resizebox{\columnwidth}{!}{
\begin{tabular}{@{}lccc@{}}
\toprule
\textbf{Model} & \textbf{Precision $\uparrow$} & \textbf{Recall $\uparrow$} & \textbf{F1 Score $\uparrow$} \\
\midrule

OWLv2\textsubscript{B}~\cite{mindererScalingOpenVocabularyObject} 
 & 0.0344
 & \underline{0.3436}
 & 0.0625
 \\

OWLv2\textsubscript{B-E}~\cite{mindererScalingOpenVocabularyObject} 
 & 0.0156
 & 0.0007
 & 0.0015
 \\
  OWLv2\textsubscript{B-F}~\cite{mindererScalingOpenVocabularyObject} 
 & 0.1078
 & 0.1801
 & 0.1349
 \\
OWLv2\textsubscript{L}~\cite{mindererScalingOpenVocabularyObject} 
 & \underline{0.3408}
 & 0.2746
 & \textbf{0.3041}
 \\
OWLv2\textsubscript{L-E}~\cite{mindererScalingOpenVocabularyObject} 
 & 0.1804
 & \textbf{0.4388}
 & \underline{0.2557}
 \\
 OWLv2\textsubscript{L-F}~\cite{mindererScalingOpenVocabularyObject}  & -- & -- & -- \\
OmDet-Turbo\textsubscript{T}~\cite{zhao2024real} 
 & 0.0082
 & 0.3142
 & 0.0161 \\
Grounding-DINO\textsubscript{T}~\cite{liuGroundingDINOMarrying2025} 
 & 0.0311
 & 0.2484
 & 0.0554
 \\
Grounding-DINO\textsubscript{B}~\cite{liuGroundingDINOMarrying2025} 
 & \textbf{0.6774}
 & 0.0166
 & 0.0325
 \\
OWL-ViT\textsubscript{B16}~\cite{mindererSimpleOpenVocabularyObject} 
 & -- & -- & -- \\
OWL-ViT\textsubscript{B32}~\cite{mindererSimpleOpenVocabularyObject}  & -- & -- & -- \\
OWL-ViT\textsubscript{L}~\cite{mindererSimpleOpenVocabularyObject} 
 & 0.1537
 & 0.2523
 & 0.1911
 \\
\bottomrule
\end{tabular}
}
\caption{Evaluation of open-vocabulary models for the task of VRU detection, with \textbf{best} and \underline{second-best} results highlighted. Model sizes, patch sizes, and the usage of ensembles and fine-tuning are indicated in subscript.}
\label{tab:open_vocab_eval}
\end{table}

For a diverse selection, it is desirable to include a representative of each model architecture. For detections that include the classes of interest, a high \textit{recall} is of interest, as it demonstrates a high true-positive ratio. While high \textit{precision} is valuable for singular models, as it reduces false positives, it is of less importance in an ensemble, where model consensus reduces noise. While three models could not detect VRUs, the OWLv2~\cite{mindererScalingOpenVocabularyObject} family performs comparably well. Thus, the final model ensemble, as depicted in Fig.~\ref{fig:data_selection}, consists of the five models OWLv2\textsubscript{B}, OWLv2\textsubscript{L-E}, OmDet-Turbo\textsubscript{T}, Grounding-DINO\textsubscript{T}, and OWL-ViT\textsubscript{L}. Identifying VRU instances requires a $3/5$ majority consensus.

\begin{figure*}[t]
  \centering
  \resizebox{\textwidth}{!}{
  \input{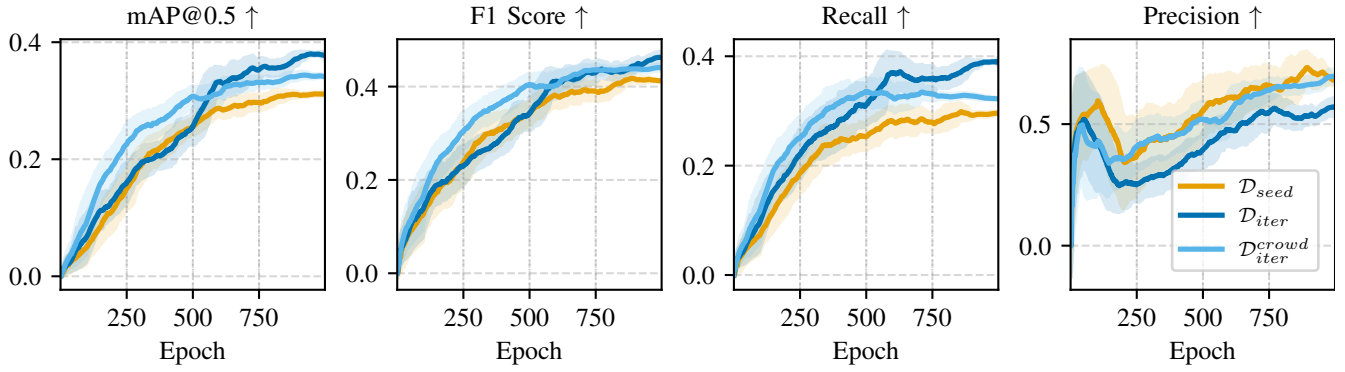}
  }
  \caption{\textbf{Model Training:} Iterative Model Improvement of a YOLOv11\textsubscript{N}~\cite{yolo11_ultralytics} model for VRU detection through an iteration of data selection with the Mcity Data Engine. The orange line shows training on the original seed dataset $\mathcal{D}_{seed}$, and the blue lines show training runs on $\mathcal{D}_{iter}$, a dataset with newly labeled VRUs, and $\mathcal{D}_{iter}^{crowd}$, with additional VRUs in crowds.}
  \label{fig:model_training}
\end{figure*}

\subsection{Data Selection and Labeling}

As the data domain of the SIP is vehicle-focused, the ensemble-based selection process eliminated 56.54\% of incoming data immediately, as no VRUs were present. Selecting only frames with multiple instances of VRUs eliminated 99.34\% of the unlabeled data stream, which demonstrates the effectiveness of the Mcity Data Engine. This allows MDE users to focus on the relevant samples from a large amount of unlabeled data, which would otherwise be infeasible as manual inspection of tens of thousands of images is a time-intensive and costly process. 

\begin{figure}[b!]
    \centering
    \subfloat[]{\includegraphics[width=0.495\columnwidth]{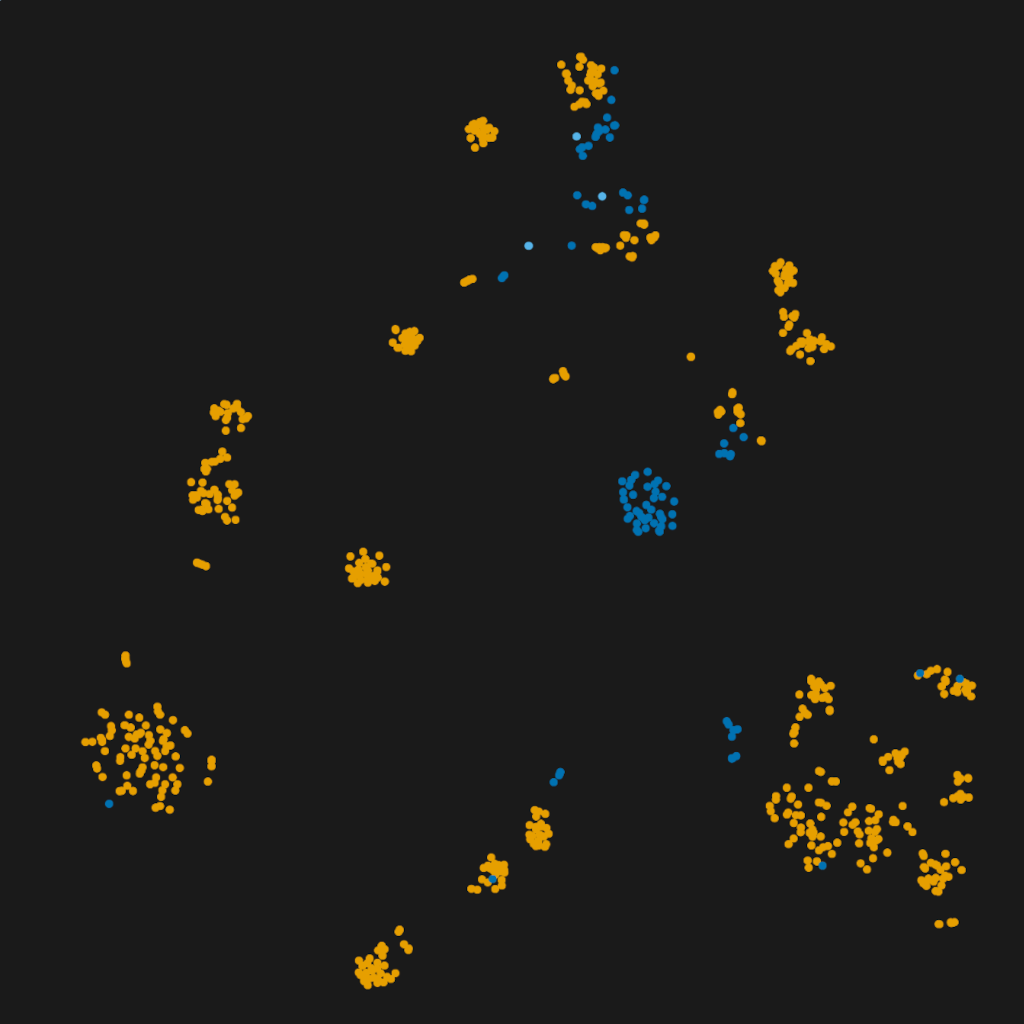}} 
    \hfill 
    \subfloat[]{\includegraphics[width=0.495\columnwidth]{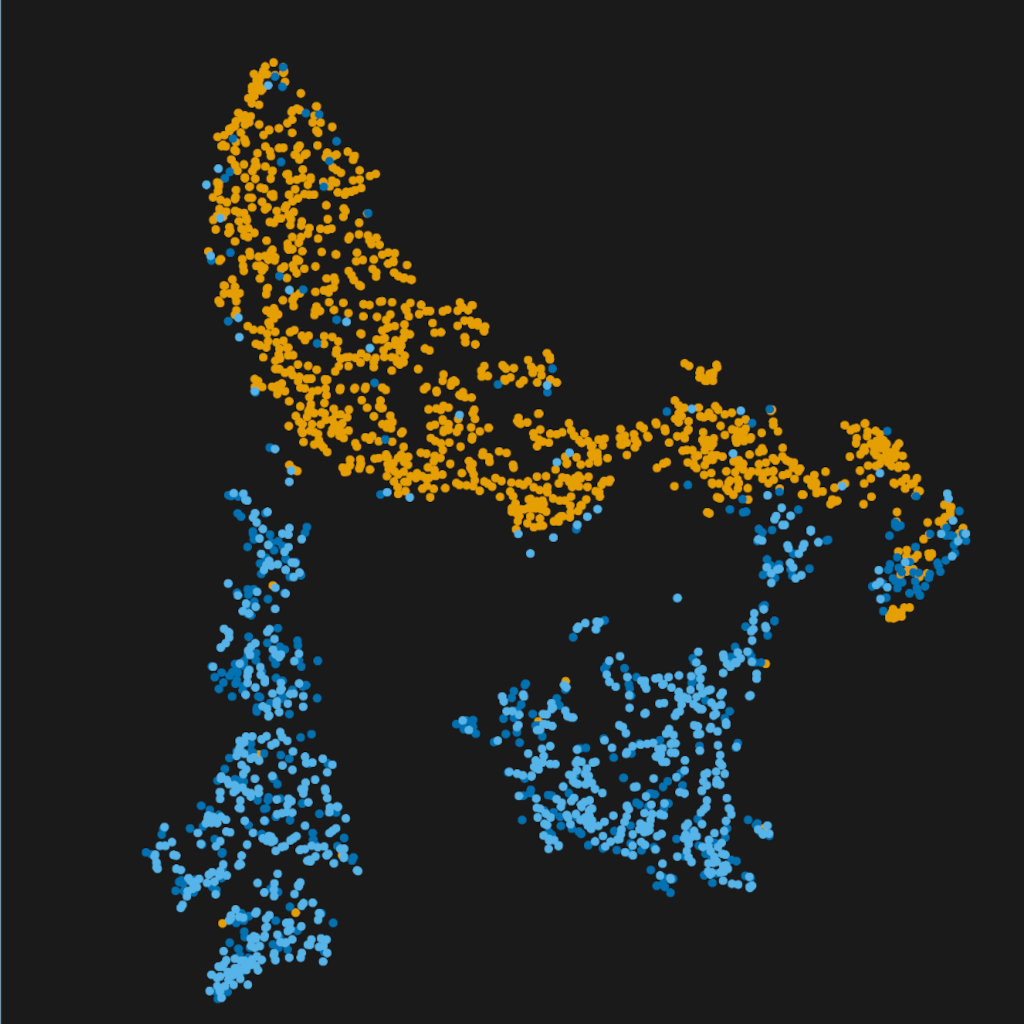}}
    \caption{Data embeddings~\cite{moore2020fiftyone} of whole frames (a) and individual VRUs (b): Orange dots represent data from $\mathcal{D}_{seed}$. Blue dots represent newly labeled data. Adding dark blue dots to $\mathcal{D}_{seed}$ forms $\mathcal{D}_{iter}$; adding all blue dots forms  $\mathcal{D}_{iter}^{crowd}$, with light blue dots representing VRUs in crowds.}
    \label{fig:data_embeddings}
\end{figure}

\begin{table}[b!]
\resizebox{\columnwidth}{!}{%
\begin{tabular}{@{}llcccc@{}}
\toprule
 \textbf{Selection}                              & \textbf{Dataset}               & \textbf{mAP@0.5} & \textbf{F1 Score} & \textbf{Precision} & \textbf{Recall} \\ \midrule
\multirow{2}{*}{\begin{tabular}[c]{@{}l@{}}Best\\mAP@0.5\end{tabular}}   & $\mathcal{D}_{iter}$        & \textbf{17.45\%}             &       -0.22\%         & 4.17\%                & -2.45\%             \\
                               & $\mathcal{D}_{iter}^{crowd}$ & 6.91\%             & -1.58\%               & -4.46\%                & 0.02\%             \\ \midrule
\multirow{2}{*}{\begin{tabular}[c]{@{}l@{}}Best\\F1 Score\end{tabular}} & $\mathcal{D}_{iter}$        & 12.37\%             & \textbf{6.39\%}               & -0.84\%                & 10.76\%             \\ 
                               & $\mathcal{D}_{iter}^{crowd}$ & 0.79\%             & 2.47\%               & \underline{11.58\%}                & -1.88\%             \\ \midrule
                               \multirow{2}{*}{\begin{tabular}[c]{@{}l@{}}Best\\Recall\end{tabular}}   & $\mathcal{D}_{iter}$        & -9.90\%             &       -12.35\%         & -39.16\%                & \textbf{26.05\%}             \\
                               & $\mathcal{D}_{iter}^{crowd}$ & -16.98\%             & -13.12\%               & -35.54\%                & 14.45\%             \\ \midrule
                               \multirow{2}{*}{\begin{tabular}[c]{@{}l@{}}Best\\Precision\end{tabular}}   & $\mathcal{D}_{iter}$        & 14.95\%             &       -0.66\%         & -8.37\%                & 2.37\%             \\
                               & $\mathcal{D}_{iter}^{crowd}$ & -25.18\%             & -21.17\%               & -10.60\%                & 0.02\%             \\ \midrule
                               \multirow{2}{*}{\begin{tabular}[c]{@{}l@{}}mAP@0.5 \&\\F1 Score\end{tabular}}   & $\mathcal{D}_{iter}$        & 14.50\%             &       \underline{5.31\%}         & -6.08\%                & 12.79\%             \\
                               & $\mathcal{D}_{iter}^{crowd}$ & 6.91\%             & 2.31\%               & \textbf{13.58\%}                & -2.94\%             \\ \midrule
                               \multirow{2}{*}{\begin{tabular}[c]{@{}l@{}}mAP@0.5 \&\\Recall\end{tabular}}   & $\mathcal{D}_{iter}$        & \underline{15.03\%}             &       -8.03\%         & -22.58\%                & \underline{19.05\%}             \\
                               & $\mathcal{D}_{iter}^{crowd}$ & 0.94\%             & -17.56\%               & -23.23\%                & 6.71\%             \\ \bottomrule
\end{tabular}%
}
\caption{\textbf{Model Validation:} Percentage metric changes based on different selection strategies, with \textbf{best} and \underline{second-best} improvements highlighted.}
\label{tab:eval_results}
\end{table}

Based on this process, a selection of 100 frames was labeled in-house with the CVAT integration and subsequently merged with the original seed dataset $\mathcal{D}_{seed}$. The resulting dataset $\mathcal{D}_{iter}^{crowd}$ contains 2,352 labeled VRU instances. The new labels represent an increase in labeled VRUs of 106.13\%. When a frame contains over 40 VRU instances, they are considered part of a \textit{crowd}, which is the case for 772 VRUs in total. This information is added as metadata without changing their labels. We refer to the new dataset without crowd VRUs as $\mathcal{D}_{iter}$, as crowd detection is considered a different task~\cite{ranasinghe2024crowddiff}, and we are interested in the effect of adding crowd VRUs to the data. The distribution of both the original data $\mathcal{D}_{seed}$ and the newly added data is visualized in Fig.~\ref{fig:data_embeddings}, showing an extension of the ODD through cameras at new locations and visually different VRUs.

\subsection{Iterative Model Improvement and Validation}

To use the newly acquired data for iterative model improvement, we select 20\% of the total data for validation and train a YOLOv11\textsubscript{N}~\cite{yolo11_ultralytics} object detection model with three different training datasets for 1,000 epochs each. First, we only use the original $\mathcal{D}_{seed}$ dataset. Then, we add the newly labeled data -- acquired through one iteration of data selection with the MDE -- to form $\mathcal{D}_{iter}$, without VRUs in crowds. Finally, we also include crowd data to form $\mathcal{D}_{iter}^{crowd}$. The results on the validation set during training are visualized in Fig.~\ref{fig:model_training}, where rolling average and standard deviation are shown. Which metric to prioritize for model selection for deployment depends on the use case at hand. While mAP@0.5 is the most comprehensive metric for overall model performance over multiple confidence thresholds, the remaining metrics evaluate the model performance based on a single confidence threshold, which is closer to a deployment setting. The F1 score represents the balance between recall and precision. High recall reduces false negatives, while high precision reduces false positives. 

Table~\ref{tab:eval_results} shows percentual improvements based on the prioritized metrics, including geometric means of multiple metrics. This model validation allows for an informed model selection for deployment. Based on the mAP@0.5 metric, the best results are achieved when crowd data is not included. Through a single iteration with the Mcity Data Engine, we can observe a significant improvement in mAP@0.5 of 17.45\%, but this comes with slightly reduced recall. Training with the complete $\mathcal{D}_{iter}^{crowd}$ dataset shows an improvement in mAP@0.5 of only 6.91\% but maintains recall compared to the model trained on $\mathcal{D}_{seed}$. If as few VRUs as possible should be missed, recall can be improved by up to 26.05\%. Selecting model weights based on the geometric mean of both mAP@0.5 and F1 score can lead to significant improvements across most metrics. For the challenging use case of VRU detection in traffic data, we are interested in missing as few pedestrians as possible while not triggering too many false alarms. This suggests choosing model weights based on the $\mathcal{D}_{iter}$ training one at the highest geometric mean of mAP@0.5 and F1 score. Here, we see strong improvements in both mAP@0.5 and recall, with a slightly increased F1 score and a slight decrease in precision. Selecting model weights at the best recall value or at the best geometric mean of mAP@0.5 and recall leads to a drastic increase in false positives. The selected model weights can now be deployed in the cloud or exported and deployed on the edge.
\section{Conclusion}
\label{sec:concl}

In this work, we have first provided an overview of current data engines and have highlighted a lack of coverage and availability among academic data engines, as shown in Table~\ref{tab:sota}. Addressing this gap, the open-source Mcity Data Engine provides all modules necessary for the complete data-based development cycle, similar to full-fledged industrial proprietary solutions, as shown in Fig.~\ref{fig:mcity_data_engine}. This enables researchers and the open-source community to leverage existing data sources and iteratively improve models without the need to develop their own solutions while remaining flexible enough to extend and adapt the MDE based on their needs.

By selecting data of interest with a robust ensemble-based open-vocabulary approach, rare and novel classes can be found in an efficient manner within large amounts of unlabeled data through the use of natural language, making the MDE accessible to non-experts as shown in Fig.~\ref{fig:data_selection}.

For future work, we are interested in leveraging a wider selection of VLMs for data selection based on natural language. VLMs with improved prediction quality might reduce the need for a computationally intensive ensemble. In addition, we would like to evaluate our approach with more iterations of data labeling and model training to further improve the observed precision-recall tradeoff. Finally, integrating data types beyond 2D images into the MDE is of interest to us.
\section*{Acknowledgment}
\label{sec:ackno}

This work was made possible through a scholarship provided by Mcity, funding Daniel Bogdoll as a visiting research scholar. The stay was financially supported by the KIT Research Travel Grant and the KIT Graduate School UpGrade Mobility.

\bibliographystyle{IEEEtran}
\bibliography{references}

\end{document}